\documentclass{article}
\usepackage[margin=1in]{geometry}
\usepackage{amsmath,amsthm,amssymb}
\usepackage{algorithm,algorithmic}
\usepackage{graphicx}
\usepackage{xcolor}
\usepackage{natbib}
\usepackage{hyperref}

\begin{document}
\newcommand{\R}{\mathbb{R}}
\newcommand{\argmax}{\mathop{\mathrm{argmax}}}
\newcommand{\Normal}{\mathrm{Normal}}
\newcommand{\EI}{\mathrm{EI}}
\newcommand{\KG}{\mathrm{KG}}
\newcommand{\ES}{\mathrm{ES}}
\newcommand{\PES}{\mathrm{PES}}
\newcommand{\xstar}{x^*}

\title{A Tutorial on Bayesian Optimization}
\author{Peter I. Frazier}
\maketitle
\abstract{
Bayesian optimization is an approach to optimizing objective functions that take a long time (minutes or hours) to evaluate.  It is best-suited for optimization over continuous domains of less than 20 dimensions, and tolerates stochastic noise in function evaluations.  It builds a surrogate for the objective and quantifies the uncertainty in that surrogate using a Bayesian machine learning technique, Gaussian process regression, and then uses an acquisition function defined from this surrogate to decide where to sample.  In this tutorial, we describe how Bayesian optimization works, including Gaussian process regression and three common acquisition functions: expected improvement, entropy search, and knowledge gradient.  We then discuss more advanced techniques, including running multiple function evaluations in parallel, multi-fidelity and multi-information source optimization, expensive-to-evaluate constraints, random environmental conditions, multi-task Bayesian optimization, and the inclusion of derivative information.  We conclude with a discussion of Bayesian optimization software and future research directions in the field.  Within our tutorial material we provide a generalization of expected improvement to noisy evaluations, beyond the noise-free setting where it is more commonly applied.  This generalization is justified by a formal decision-theoretic argument, standing in contrast to previous ad hoc modifications.
}
\vskip 2mm

\section{Introduction}
\label{sec:intro}
Bayesian optimization (BayesOpt) is a class of machine-learning-based optimization methods focused on solving the problem
\begin{equation}
\max_{x \in A} f(x), \label{eq:obj}
\end{equation}
where the feasible set and objective function typically have the following properties:
\begin{itemize}
\item The input $x$ is in $\R^d$ for a value of $d$ that is not too large.  Typically $d\le 20$ in most successful applications of BayesOpt.
\item The feasible set $A$ is a simple set, in which it is easy to assess membership.  Typically $A$ is a hyper-rectangle $\{ x \in \R^d : a_i \le x_i \le b_i \}$ or the $d$-dimensional simplex $\{ x \in \R^d : \sum_i x_i = 1 \}$.  Later (Section~\ref{sec:exotic}) we will relax this assumption.
\item The objective function $f$ is continuous.  This will typically be required to model $f$ using Gaussian process regression.
\item $f$ is ``expensive to evaluate'' in the sense that the number of evaluations that may be performed is limited, typically to a few hundred.  This limitation typically arises because each evaluation takes a substantial amount of time (typically hours), but may also occur because each evaluation bears a monetary cost (e.g., from purchasing cloud computing power, or buying laboratory materials), or an opportunity cost (e.g., if evaluating $f$ requires asking a human subject questions who will tolerate only a limited number). 

\item $f$ lacks known special structure like concavity or linearity that would make it easy to optimize using techniques that leverage such structure to improve efficiency.  We summarize this by saying $f$ is a ``black box.''
\item When we evaluate $f$, we observe only $f(x)$ and no first- or second-order derivatives.  This prevents the application of first- and second-order methods like gradient descent, Newton's method, or quasi-Newton methods.  We refer to problems with this property as ``derivative-free''.
\item Through most of the article, we will assume $f(x)$ is observed without noise. Later (Section~\ref{sec:exotic}) we will allow $f(x)$ to be obscured by stochastic noise.  In almost all work on Bayesian optimization, noise is assumed independent across evaluations and Gaussian with constant variance. 
\item Our focus is on finding a {\it global} rather than local optimum.
\end{itemize}
We summarize these problem characteristics by saying that BayesOpt is designed for black-box derivative-free global optimization.

The ability to optimize expensive black-box derivative-free functions makes BayesOpt extremely versatile.  Recently it has become extremely popular for tuning hyperparameters in machine learning algorithms, especially deep neural networks \citep{snoek2012practical}.
Over a longer period, since the 1960s, BayesOpt has been used extensively for designing engineering systems \citep{Mockus89,JoScWe98,FoSoKe08}.  BayesOpt has also been used to choose laboratory experiments in materials and drug design \citep{FrazierNegoescuPowell2011,frazier2016bayesian,packwood2017bayesian}, in calibration of environmental models \citep{ShReFl07}, and in reinforcement learning \citep{BrCoFr09,Li08,LiWaBoSc07}. 

BayesOpt originated with the work of Kushner \citep{Kushner64}, Zilinskas \citep{Zi75,MoTiZi78}, and Mo\v{c}kus \citep{mockus1975bayesian,Mockus89}, but received substantially more attention after that work was popularized by \cite{JoScWe98} and their work on the Efficient Global Optimization (EGO) algorithm.  Following \cite{JoScWe98}, innovations developed in that same literature include multi-fidelity optimization \citep{HuAlNoMi06,SobesterLearyKeane2004}, multi-objective optimization \citep{Ke06,Kn06,MoMo91}, and a study of convergence rates \citep{Ca97,CaZi00,CaZi05,CaZi99}.  The observation made by \cite{snoek2012practical} that BayesOpt is useful for training deep neural networks sparked a surge of interest within machine learning, with complementary innovations from that literature including multi-task optimization \citep{swersky2013multi,toscano2018bayesian}, multi-fidelity optimization specifically aimed at training deep neural networks \citep{klein2016fast}, and parallel methods \citep{GiLeCa08,GiLeCa10,wang2016parallel,wu2016parallel}.  Gaussian process regression, its close cousin kriging, and BayesOpt have also been studied recently in the simulation literature \citep{kleijnen2008design,salemi2014discrete,mehdad2018efficient} for modeling and optimizing systems simulated using discrete event simulation.

There are other techniques outside of BayesOpt that can be used to optimize expensive derivative-free black-box functions.  While we do not review methods from this literature here in detail, many of them have a similar flavor to BayesOpt methods: they maintain a surrogate that models the objective function, which they use to choose where to evaluate \citep{BoDeFrSe99,RegisShoemaker2007,ReSh07,ReSh05}. This more general class of methods is often called ``surrogate methods.''
Bayesian optimization distinguishes itself from other surrogate methods by using surrogates developed using Bayesian statistics, and in deciding where to evaluate the objective using a Bayesian interpretation of these surrogates.  

We first introduce the typical form that Bayesian optimization algorithms take in Section~\ref{sec:overview}.  This form involves two primary components: a method for statistical inference, typically Gaussian process (GP) regression; and an acquisition function for deciding where to sample, which is often expected improvement.  We describe these two components in detail in Sections~\ref{sec:GP} and \ref{sec:EI}.  
We then describe three alternate acquisition functions: knowledge-gradient (Section~\ref{sec:KG}), entropy search, and predictive entropy search (Section~\ref{sec:EntropySearch}). 
These alternate acquisition functions are particularly useful in problems falling outside the strict set of assumptions above, which we call ``exotic'' Bayesian optimization problems and we discuss in Section~\ref{sec:exotic}.
These exotic Bayesian optimization problems include those with parallel evaluations, constraints, multi-fidelity evaluations, multiple information sources, random environmental conditions, multi-task objectives, and derivative observations.  
We then discuss Bayesian optimization and Gaussian process regression software in Section~\ref{sec:software} and conclude with a discussion of future research directions in Section~\ref{sec:conclusion}.

Other tutorials and surveys on Bayesian optimization include \cite{shahriari2016taking,BrCoFr09,Sa02,frazier2016bayesian}.
This tutorial differs from these others in its coverage of non-standard or ``exotic'' Bayesian optimization problems.  It also differs in its substantial emphasis on acquisition functions, with less emphasis on GP regression.  Finally, it includes what we believe is a novel analysis of expected improvement for noisy measurements, and argues that the acquisition function previously proposed by \cite{ScottFrazierPowell2011} is the most natural way to apply the expected improvement acquisition function when measurements are noisy.

\section{Overview of BayesOpt}
\label{sec:overview}

BayesOpt consists of two main components: a Bayesian statistical model for modeling the objective function,
and an acquisition function for deciding where to sample next.
After evaluating the objective according to an initial space-filling experimental design, often consisting of points chosen uniformly at random, they are used iteratively to allocate the remainder of a budget of $N$ function evaluations, as shown in Algorithm~\ref{alg:BayesOpt}.

\begin{algorithmic}
\begin{algorithm}
\STATE Place a Gaussian process prior on $f$
\STATE Observe $f$ at $n_0$ points according to an initial space-filling experimental design.  Set $n=n_0$.
\WHILE{$n \le N$}
\STATE Update the posterior probability distribution on $f$ using all available data
\STATE Let $x_n$ be a maximizer of the acquisition function over $x$, where the acquisition function is computed using the current posterior distribution.
\STATE Observe $y_n = f(x_n)$.
\STATE Increment $n$
\ENDWHILE
\STATE Return a solution: either the point evaluated with the largest $f(x)$, or the point with the largest posterior mean.
\caption{
Basic pseudo-code for Bayesian optimization
\label{alg:BayesOpt}
}
\end{algorithm}
\end{algorithmic}

The statistical model, which is invariably a Gaussian process, provides a Bayesian posterior probability distribution
 that describes potential values for $f(x)$ at a candidate point $x$.  Each time we observe $f$ at a new point, this posterior distribution is updated.  We discuss Bayesian statistical modeling using Gaussian processes in detail in Section~\ref{sec:GP}.
The acquisition function measures the value that would be generated by evaluation of the objective function at a new point $x$, based on the current posterior distribution over $f$.  
We discuss expected improvement, the most commonly used acquisition function, in Section~\ref{sec:EI}, and then discuss other acquisition functions in Section~\ref{sec:KG} and \ref{sec:EntropySearch}.

\begin{figure}[tb]
\centering
\includegraphics[width=0.8\textwidth]{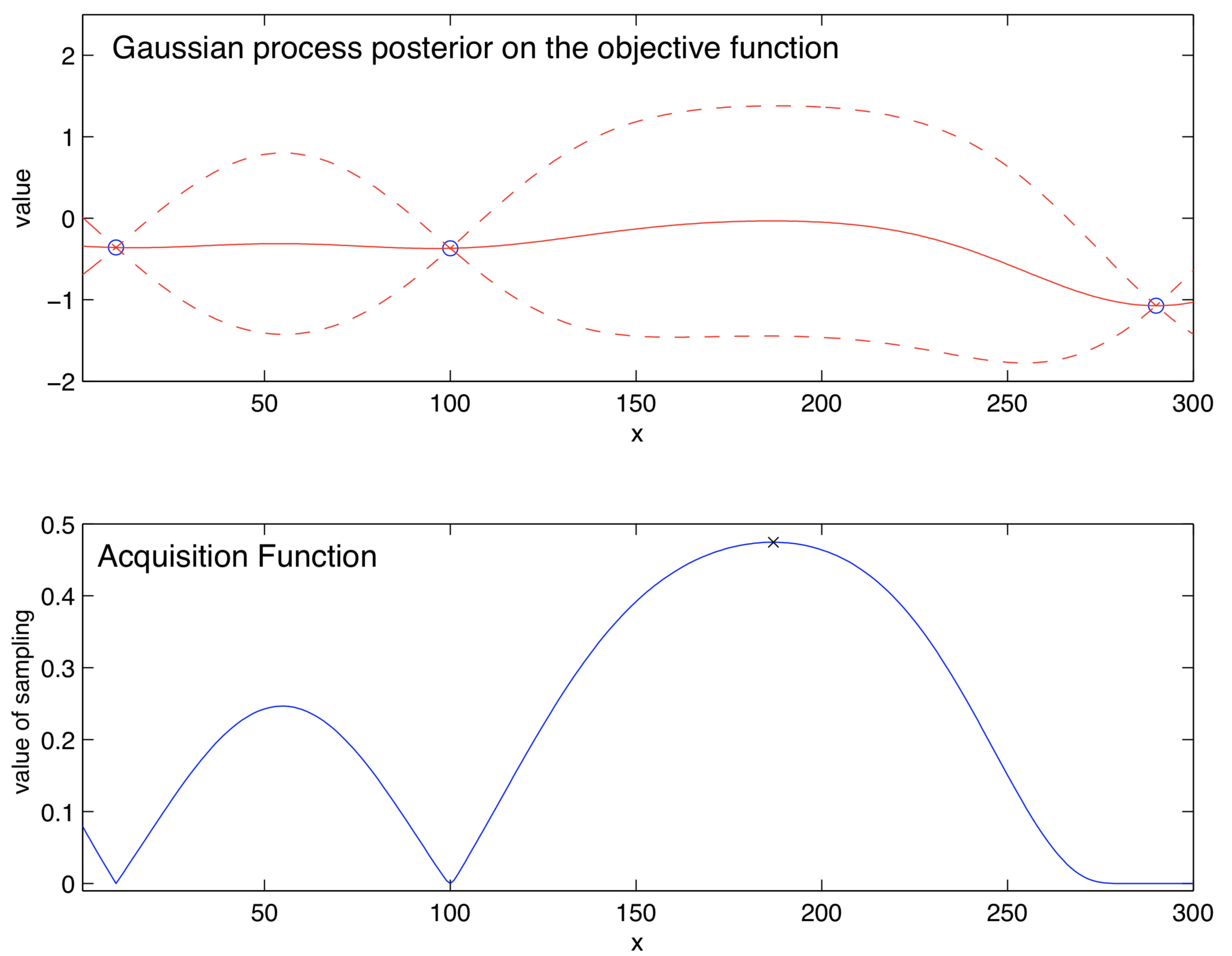}
\caption{Illustration of BayesOpt, maximizing an objective function $f$ with a 1-dimensional continuous input.  The top panel shows: noise-free observations of the objective function $f$ at 3 points, in blue; an estimate of $f(x)$ (solid red line); and Bayesian credible intervals (similar to confidence intervals) for $f(x)$ (dashed red line).  These estimates and credible intervals are obtained using GP regression.  The bottom panel shows the acquisition function.  Bayesian optimization chooses to sample next at the point that maximizes the acquisition function, indicated here with an ``x.''
\label{fig:BayesOptExample}
}
\end{figure}

One iteration of BayesOpt from Algorithm~\ref{alg:BayesOpt} using GP regression and expected improvement is illustrated in Figure~\ref{fig:BayesOptExample}.
The top panel shows noise-free observations of the objective function with blue circles at three points.  It also shows the output of GP regression. 
We will see below in Section~\ref{sec:GP} that GP regression produces a posterior probability distribution on each $f(x)$ that is normally distributed with mean $\mu_n(x)$ and variance $\sigma^2_n(x)$.
This is pictured in the figure with $\mu_n(x)$ as the solid red line,  and a 95\% Bayesian credible interval for $f(x)$, $\mu_n(x) \pm 1.96\times\sigma_n(x)$, as dashed red lines.  
The mean can be interpreted as a point estimate of $f(x)$. The credible interval acts like a confidence interval in frequentist statistics, and contains $f(x)$ with probability 95\% according to the posterior distribution.  The mean interpolates the previously evaluated points. The credible interval has 0 width at these points, and grows wider as we move away from them.

The bottom panel shows the expected improvement acquisition function that corresponds to this posterior.  Observe that it takes value 0 at points that have previously been evaluated.  This is reasonable when evaluations of the objective are noise-free because evaluating these points provides no useful information toward solving \eqref{eq:obj}.  Also observe that it tends to be larger for points with larger credible intervals, because observing a point where we are more uncertain about the objective tends to be more useful in finding good approximate global optima.  Also observe it tends to be larger for points with larger posterior means, because such points tend to be near good approximate global optima.

We now discuss the components of BayesOpt in detail, first discussing GP regression in Section~\ref{sec:GP}, then discuss acquisition functions in Section~\ref{sec:acquisition}, starting with expected improvement in Section~\ref{sec:EI}.  We then discuss more sophisticated acquisition functions (knowledge gradient, entropy search, and predictive entropy search) in Sections~\ref{sec:KG} and \ref{sec:EntropySearch}.  Finally, we discuss extensions of the basic problem described in Section~\ref{sec:intro} in Section~\ref{sec:exotic}, discussing problems with measurement noise, parallel function evaluations, constraints, multi-fidelity observations, and others.

\section{Gaussian Process (GP) Regression}
\label{sec:GP}

GP regression is a Bayesian statistical approach for modeling functions.  We offer a brief introduction here.  A more complete treatment may be found in \cite{RaWi06}.

We first describe GP regression, focusing on $f$'s values at a finite collection of points $x_1,\ldots,x_k \in \R^d$.  It is convenient to collect the function's values at these points together into a vector $[f(x_1),\ldots,f(x_k)]$.
Whenever we have a quantity that is unknown in Bayesian statistics, like this vector, we suppose that it was drawn at random by nature from some prior probability distribution.  
GP regression takes this prior distribution to be multivariate normal, with a particular mean vector and covariance matrix.

We construct the mean vector by evaluating a {\it mean function} $\mu_0$ at each $x_i$.  
We construct the covariance matrix by evaluating a {\it covariance function} or {\it kernel} $\Sigma_0$ at each pair of points $x_i$, $x_j$.  The kernel is chosen so that points $x_i,x_j$ that are closer in the input space have a large positive correlation, encoding the belief that they should have more similar function values than points that are far apart.  The kernel should also have the property that the resulting covariance matrix is positive semi-definite, regardless of the collection of points chosen.  
Example mean functions and kernels are discussed below in Section~\ref{sec:prior}.

The resulting prior distribution on $[f(x_1),\ldots,f(x_k)]$ is,
\begin{equation}
f(x_{1:k}) \sim \Normal\left(\mu_0(x_{1:k}),\Sigma_0(x_{1:k},x_{1:k})\right), \label{eq:prior}
\end{equation}
where we use compact notation for functions applied to collections of input points: $x_{1:k}$
indicates the sequence $x_1,\ldots,x_k$,
$f(x_{1:k}) = [f(x_1),\ldots,f(x_k)]$, $\mu_0(x_{1:k}) = [\mu_0(x_1),\ldots,\mu_0(x_k)]$, and 
$\Sigma_0(x_{1:k},x_{1:k}) = [\Sigma_0(x_1,x_1), \ldots, \Sigma_0(x_1,x_k); \ldots; \Sigma_0(x_k,x_1), \ldots, \Sigma_0(x_k,x_k)]$.

Suppose we observe $f(x_{1:n})$ without noise for some $n$ and we wish to infer the value of $f(x)$ at some new point $x$.  To do so, we let $k=n+1$ and $x_{k} = x$, so that the prior over $[f(x_{1:n}), f(x)]$ is given by \eqref{eq:prior}.  We may then compute the conditional distribution of $f(x)$ given these observations using Bayes' rule (see details in Chapter 2.1 of \cite{RaWi06}),  
\begin{equation}
\begin{split}
\label{eq:posterior}
f(x) | f(x_{1:n}) &\sim \Normal(\mu_n(x), \sigma^2_n(x)) \\
\mu_n(x) &= \Sigma_0(x,x_{1:n}) \Sigma_0(x_{1:n},x_{1:n})^{-1} \left(f(x_{1:n}) - \mu_0(x_{1:n})\right) + \mu_0(x) \\
\sigma^2_n(x) &= 
\Sigma_0(x,x) - \Sigma_0(x,x_{1:n}) \Sigma_0(x_{1:n},x_{1:n})^{-1} \Sigma_0(x_{1:n},x).
\end{split}
\end{equation}
	
	This conditional distribution is called the {\it posterior probability distribution} in the nomenclature of Bayesian statistics.  
The posterior mean $\mu_n(x)$ is a weighted average between the prior $\mu_0(x)$ and an estimate based on the data $f(x_{1:n})$, with a weight that depends on the kernel.
The posterior variance $\sigma_n^2(x)$ is equal to the prior covariance $\Sigma_0(x,x)$ less a term that corresponds to the variance removed by observing $f(x_{1:n})$.

Rather than computing posterior means and variances directly using \eqref{eq:posterior} and matrix inversion, it is typically faster and more numerically stable to use a Cholesky decomposition and then solve a linear system of equations.  This more sophisticated technique is discussed as Algorithm~2.1 in Section 2.2 of \cite{RaWi06}.  Additionally, to improve the numerical stability of this approach or direct computation using \eqref{eq:posterior}, it is often useful to add a small positive number like $10^{-6}$ to each element of the diagonal of $\Sigma_0(x_{1:n},x_{1:n})$, especially when $x_{1:n}$ contains two or more points that are close together.  
This prevents eigenvalues of $\Sigma_0(x_{1:n},x_{1:n})$ from being too close to $0$, and only changes the predictions that would be made by an infinite-precision computation by a small amount.

Although we have modeled $f$ at only a finite number of points, the same approach can be used when modeling $f$ over a continuous domain $A$.  Formally a {\it Gaussian process} with mean function $\mu_0$ and kernel $\Sigma_0$ is a probability distribution over the function $f$ with the property that, for any given collection of points $x_{1:k}$, the marginal probability distribution on $f(x_{1:k})$ is given by \eqref{eq:prior}.  Moreover, the arguments that justified \eqref{eq:posterior} still hold when our prior probability distribution on $f$ is a Gaussian process. 

In addition to calculating the conditional distribution of $f(x)$ given $f(x_{1:n})$, it is also possible to calculate the conditional distribution of $f$ at more than one unevaluated point.  The resulting distribution is multivariate normal, with a mean vector and covariance kernel that depend on the location of the unevaluated points, the locations of the measured points $x_{1:n}$, and their measured values $f(x_{1:n})$.  The functions that give entries in this mean vector and covariance matrix have the form required for a mean function and kernel described above, 
and the conditional distribution of $f$ given $f(x_{1:n})$ is a Gaussian process with this mean function and covariance kernel.

\subsection{Choosing a Mean Function and Kernel} 
\label{sec:prior}

We now discuss the choice of kernel.  Kernels typically have the property that points closer in the input space are more strongly correlated, i.e., that if $||x-x'|| < ||x-x''||$ for some norm $||\cdot||$, then $\Sigma_0(x,x') > \Sigma_0(x,x'')$.  Additionally, kernels are required to be positive semi-definite functions.
Here we describe two example kernels and how they are used.

\begin{figure}[tb]
\centering
\includegraphics[width=0.8\textwidth]{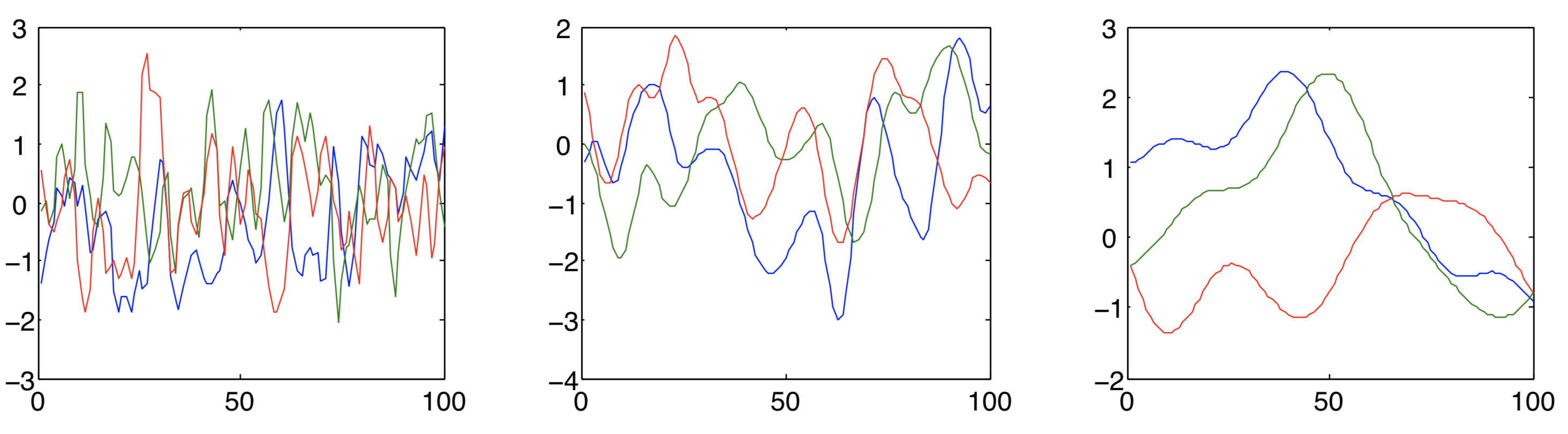}
\caption{Random functions $f$ drawn from a Gaussian process prior with a power exponential kernel.  Each plot corresponds to a different value for the parameter $\alpha_1$, with $\alpha_1$ decreasing from left to right.  Varying this parameter creates different beliefs about how quickly $f(x)$ changes with $x$.
\label{fig:vary_hypers}}
\end{figure}

One commonly used and simple kernel is the {\it power exponential} or {\it Gaussian} kernel,
\begin{equation*}
\Sigma_0(x,x') = \alpha_0 \exp\left( - ||x-x'||^2 \right),
\end{equation*}
where $||x-x'||^2 = \sum_{i=1}^d \alpha_i (x_i-x'_i)^2$, and $\alpha_{0:d}$ are parameters of the kernel.  
Figure~\ref{fig:vary_hypers} shows random functions with a 1-dimensional input drawn from a Gaussian process prior with a power exponential kernel with different values of $\alpha_1$.  Varying this parameter creates different beliefs about how quickly $f(x)$ changes with $x$. 

Another commonly used kernel is the {\it M\`atern} kernel,
\begin{equation*}
\Sigma_0(x,x') = \alpha_0 \frac{2^{1-\nu}}{\Gamma(\nu)} \left(\sqrt{2\nu} || x-x' ||\right)^\nu K_\nu(\sqrt{2\nu} || x-x' ||) 
\end{equation*}
where $K_\nu$ is the modified Bessel function, and we have a parameter $\nu$ in addition to the parameters $\alpha_{0:d}$.
We discuss choosing these parameters below in Section~\ref{sec:hyperparameters}.

Perhaps the most common choice for the mean function is a constant value, $\mu_0(x) = \mu$.  When $f$ is believed to have a trend or some application-specific parametric structure, we may also take the mean function to be 
\begin{equation}
\mu_0(x) = \mu + \sum_{i=1}^p \beta_i \Psi_i(x),
\end{equation}
where each $\Psi_i$ is a parametric function, and often a low-order polynomial in $x$. 

\subsection{Choosing Hyperparameters} 
\label{sec:hyperparameters}

The mean function and kernel contain parameters.  We typically call these parameters of the prior {\it hyperparameters}.  We indicate them via a vector $\eta$.  For example, if we use a M\`atern kernel and a constant mean function, $\eta = (\alpha_{0:d}, \nu, \mu)$.

To choose the hyperparameters, three approaches are typically considered.  The first is to find the {\it maximum likelihood estimate} (MLE).  In this approach, when given observations $f(x_{1:n})$, we calculate the likelihood of these observations under the prior, $P(f(x_{1:n}) | \eta)$, where we modify our notation to indicate its dependence on $\eta$.  This likelihood is a multivariate normal density.  Then, in maximum likelihood estimation, we set $\eta$ to the value that maximizes this likelihood,
\begin{equation*}
\hat{\eta} = \argmax_\eta P(f(x_{1:n}) | \eta)
\end{equation*}

The second approach amends this first approach by imagining that the hyperparameters $\eta$ were themselves chosen from a prior, $P(\eta)$.  
We then estimate $\eta$ by
the {\it maximum a posteriori} (MAP) estimate \citep{gelman2014bayesian}, which is the value of $\eta$ that maximizes the posterior,
\begin{equation*}
\hat{\eta} = \argmax_\eta P(\eta | f(x_{1:n})) = \argmax_\eta P(f(x_{1:n}) | \eta) P(\eta)
\end{equation*}
In moving from the first expression to the second we have used Bayes' rule and then dropped a normalization constant 
$\int P(f(x_{1:n}) | \eta') P(\eta')\, d\eta'$ that does not depend on the quantity $\eta$ being optimized.

The MLE is a special case of the MAP if we take the prior on the hyperparameters $P(\eta)$ to be the (possibly degenerate) probability distribution that has constant density over the domain of $\eta$.
The MAP is useful if the MLE sometimes estimates unreasonable hyperparameter values, for example, corresponding to functions that vary too quickly or too slowly (see Figure~\ref{fig:vary_hypers}).
By choosing a prior that puts more weight on hyperparameter values that are reasonable for a particular problem, MAP estimates can better correspond to the application.  Common choices for the prior include the uniform distribution (for preventing estimates from falling outside of some pre-specified range), the normal distribution (for suggesting that the estimates fall near some nominal value without setting a hard cutoff), and the log-normal and truncated normal distributions (for providing a similar suggestion for positive parameters).

The third approach is called the {\it fully Bayesian} approach.  In this approach, we wish to compute the posterior distribution on $f(x)$ marginalizing over all possible values of the hyperparameters,
\begin{equation}
P(f(x) = y | f(x_{1:n})) = \int P(f(x) = y | f(x_{1:n}), \eta) P(\eta | f(x_{1:n}))\, d\eta
\label{eq:fully-Bayesian}
\end{equation}
This integral is typically intractable, but we can approximate it through sampling:
\begin{equation}
P(f(x) = y | f(x_{1:n})) \approx \frac1J \sum_{j=1}^J P(f(x) = y | f(x_{1:n}), \eta = \hat{\eta}_j)
\end{equation}
where $(\hat{\eta}_j : j = 1,\ldots,J)$ are sampled from $P(\eta | f(x_{1:n}))$ via an MCMC method, e.g., slice sampling \citep{neal2003slice}.
MAP estimation can be seen as an approximation to fully Bayesian inference: if we approximate the posterior $P(\eta | f(x_{1:n}))$ by a point mass at the $\eta$ 
that maximizes the posterior density,
then inference with the MAP recovers \eqref{eq:fully-Bayesian}.

\section{Acquisition Functions}
\label{sec:acquisition}

Having surveyed Gaussian processes, we return to Algorithm~\ref{alg:BayesOpt} and discuss the acquisition function used in that loop.  We focus on the setting described in Section~\ref{sec:intro} with noise-free evaluations, which we call the ``standard'' problem, and then discuss noisy evaluations, parallel evaluations, derivative observations, and other ``exotic'' extensions in Section~\ref{sec:exotic}.

The most commonly used acquisition function is expected improvement, and we discuss it first, in Section~\ref{sec:EI}.  Expected improvement performs well and is easy to use.  We then discuss the knowledge gradient (Section~\ref{sec:KG}), entropy search and predictive entropy search (Section~\ref{sec:EntropySearch}) acquisition functions.  These alternate acquisition functions are most useful in exotic problems where an assumption made by expected improvement, that the primary benefit of sampling occurs through an improvement {\it at the point sampled}, is no longer true.

\subsection{Expected Improvement}
\label{sec:EI}

The expected improvement acquisition function is derived by a thought experiment.  
Suppose we are using Algorithm~\ref{alg:BayesOpt} to solve \eqref{eq:obj}, in which $x_n$ indicates the point sampled at iteration $n$ and $y_n$ indicates the observed value.
Assume that we may only return a solution that we have evaluated as our final solution to \eqref{eq:obj}.  
Also suppose for the moment that we have no evaluations left to make, and must return a solution based on those we have already performed.
Since we observe $f$ without noise, 
the optimal choice is the previously evaluated point with the largest observed value.  Let $f^*_n = \max_{m\le n} f(x_m)$ be the value of this point, where $n$ is the number of times we have evaluated $f$ thus far.

Now suppose in fact we have one additional evaluation to perform, and we can perform it anywhere.  If we evaluate at $x$, we will observe $f(x)$.  After this new evaluation, the value of the best point we have observed will either be $f(x)$ (if $f(x) \ge f^*_n$) or $f^*_n$ (if $f(x) \le f^*_n$).  The improvement in the value of the best observed point is then $f(x) - f^*_n$ if this quantity is positive, and $0$ otherwise.  We can write this improvement more compactly as $[f(x) - f^*_n]^+$, where $a^+ = \max(a,0)$ indicates the positive part.

While we would like to choose $x$ so that this improvement is large, $f(x)$ is unknown until after the evaluation.  What we can do, however, 
is to take the expected value of this improvement and choose $x$ to maximize it.
We define the {\it expected improvement} as,
\begin{equation}
\EI_n(x) := E_n\left[[f(x) - f^*_n]^+\right] \label{eq:EI-theory}
\end{equation}
Here, $E_n[\cdot] = E[\cdot | x_{1:n}, y_{1:n}]$ indicates the expectation taken under the posterior distribution given evaluations of $f$ at $x_1,\ldots x_n$.
This posterior distribution is given by \eqref{eq:posterior}: $f(x)$ given $x_{1:n},y_{1:n}$ is normally distributed with mean $\mu_n(x)$ and variance $\sigma^2_n(x)$.

The expected improvement can be evaluated in closed form using integration by parts, as described in \cite{JoScWe98} or \cite{clark1961greatest}.  The resulting expression is
\begin{equation}
\EI_n(x) = [\Delta_n(x)]^+ + \sigma_n(x) \varphi\left(\frac{\Delta_n(x)}{\sigma_n(x)}\right) - |\Delta_n(x)|\Phi\left(\frac{\Delta_n(x)}{\sigma_n(x)}\right),
\label{eq:EI-formula}
\end{equation}
where $\Delta_n(x) := \mu_n(x) - f^*_n$ is the expected difference in quality between the proposed point $x$ and the previous best.

The expected improvement algorithm then evaluates at the point with the largest expected improvement,
\begin{equation}
x_{n+1} = \argmax \EI_n(x), \label{eq:EI-max}
\end{equation}
breaking ties arbitrarily.  This algorithm was first proposed by Mo\v{c}kus \citep{mockus1975bayesian} but was popularized by \cite{JoScWe98}.  The latter article also used the name ``Efficient Global Optimization'' or EGO.

Implementations use a variety of approaches for solving \eqref{eq:EI-max}.  Unlike the objective $f$ in our original optimization problem \eqref{eq:obj}, $\EI_n(x)$ is inexpensive to evaluate and allows easy evaluation of first- and second-order derivatives.  Implementations of the expected improvement algorithm can then use a continuous first- or second-order optimization method to solve \eqref{eq:EI-max}.  For example, one technique that has worked well for the author is to calculate first derivatives and use the quasi-Newton method L-BFGS-B \citep{liu1989limited}.

\begin{figure}[tb]
\centering
\includegraphics[width=0.5\textwidth]{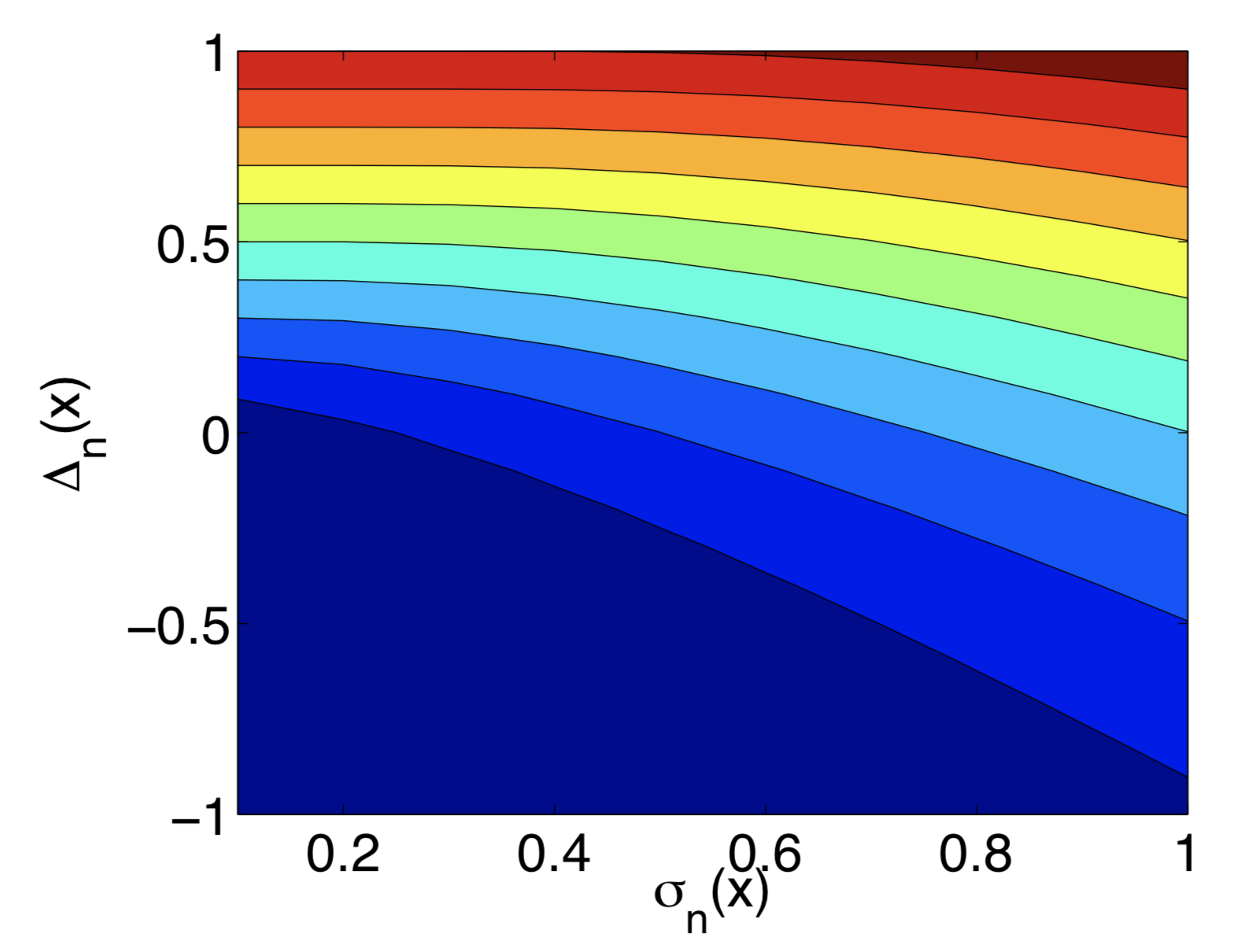}
\caption{Contour plot of $\EI(x)$, the expected improvement \eqref{eq:EI-formula}, in terms of $\Delta_n(x)$ (the expected difference in quality between the proposed point and the best previously evaluated point) and the posterior standard deviation $\sigma_n(x)$.  Blue indicates smaller values and red higher ones.  The expected improvement is increasing in both quantities, and curves of $\Delta_n(x)$ versus $\sigma_n(x)$ with equal EI define an implicit tradeoff between evaluating at points with high expected quality (high $\Delta_n(x)$ versus high uncertainty (high $\sigma_n(x)$).
\label{fig:EIcontour}}
\end{figure}

Figure~\ref{fig:EIcontour} shows the contours of $\EI_n(x)$ in terms of $\Delta_n(x)$ and the posterior standard deviation $\sigma_n(x)$.  
$\EI_n(x)$ is increasing in both $\Delta_n(x)$ and $\sigma_n(x)$.  Curves of $\Delta_n(x)$ versus $\sigma_n(x)$ with equal $EI$ show how EI balances between evaluating at points with high expected quality (high $\Delta_n(x)$)) versus high uncertainty (high $\sigma_n(x)$).  
In the context of optimization, evaluating at points with high expected quality relative to the previous best point is valuable because good approximate global optima are likely to reside at such points.
On the other hand, evaluating at points with high uncertainty is valuable because it teaches about the objective in locations where we have little knowledge and which tend to be far away from where we have previously measured.  A point that is substantially better than one we have seen previously may very well reside there.

Figure~\ref{fig:BayesOptExample} shows $\EI(x)$ in the bottom panel.  We see this tradeoff, with the largest expected improvement occurring where the posterior standard deviation is high (far away from previously evaluated points), and where the posterior mean is also high.  The smallest expected improvement is $0$, at points where we have previously evaluated.  The posterior standard deviation is $0$ at this point, and the posterior mean is necessarily no larger than the best previously evaluated point.  The expected improvement algorithm would evaluate next at the point indicated with an x, where EI is maximized.

Choosing where to evaluate based on a tradeoff between high expected performance and high uncertainty appears in other domains, including multi-armed bandits \cite{mahajan2008multi} and reinforcement learning \cite{sutton1998reinforcement}, and is often called the ``exploration vs. exploitation tradeoff'' \citep{kaelbling1996reinforcement}.

\subsection{Knowledge Gradient}
\label{sec:KG}

The knowledge-gradient acquisition function is derived by revisiting the assumption made in EI's derivation that we are only willing to return a previously evaluated point as our final solution.  That assumption is reasonable when evaluations are noise-free and we are highly risk averse, but if the decision-maker is willing to tolerate some risk then she might be willing to report a final solution that has some uncertainty attached to it.  Moreover, if evaluations have noise (discussed below in Section~\ref{sec:exotic}) then the final solution reported will necessarily have uncertain value because we can hardly evaluate it an infinite number of times. 

We replace this assumption by allowing the decision-maker to return any solution she likes, even if it has not been previously evaluated.
We also assume risk-neutrality \citep{berger2013statistical}, i.e., we value a random outcome $X$ according to its expected value.  
The solution that we would choose if we were to stop sampling after $n$ samples would be the one with the largest $\mu_n(x)$ value.  This solution (call it $\widehat{\xstar}$, since it approximates the global optimum $\xstar$) would have value $f(\widehat{\xstar})$.  $f(\widehat{\xstar})$ is random under the posterior, and has conditional expected value $\mu_n(\widehat{\xstar}) = \max_{x'} \mu_n(x') =: \mu^*_n$.

On the other hand, if we were to take one more sample at $x$, we would obtain a new posterior distribution with posterior mean $\mu_{n+1}(\cdot)$.  This posterior mean would be computed via \eqref{eq:posterior}, but including the additional observation $x_{n+1},y_{n+1}$.  If we were to report a final solution after this sample, its expected value under the new posterior distribution would be $\mu^*_{n+1} := \max_{x'} \mu_{n+1}(x')$.
Thus, the increase in conditional expected solution value due to sampling is $\mu^*_{n+1} - \mu^*_n$.

While this quantity is unknown before we sample at $x_{n+1}$, we can compute its expected value given the observations at $x_1,\ldots,x_n$ that we have.  We call this quantity the {\it knowledge gradient (KG)} for measuring at $x$,
\begin{equation}
\KG_n(x) := E_n\left[ \mu^*_{n+1} - \mu^*_n | x_{n+1} = x\right].
\label{eq:KG}
\end{equation}
Using the knowledge gradient as our acquisition function then leads us to sample at the point with largest $\KG_n(x)$, $\argmax_x \KG_n(x)$.

This algorithm was first proposed in \cite{frazier2009knowledge} for GP regression over discrete $A$, building on earlier work \citep{frazier2008knowledge} that proposed the same algorithm for Bayesian ranking and selection \citep{chick2001new} with an independent prior.  (Bayesian ranking and selection is similar to Bayesian optimization, except that $A$ is discrete and finite, observations are necessarily noisy, and the prior is typically independent across $x$.)

The conceptually simplest way to compute the KG acquisition function is via simulation, as shown in Algorithm~\ref{alg:KG}.  Within a loop, this algorithm simulates one possible value for the observation $y_{n+1}$ that may result from taking evaluation $n+1$ at a designated $x$.  Then it computes what the maximum of the new posterior mean $\mu^*_{n+1}$ would be if that value for $y_{n+1}$ were the one that actually resulted from the measurement.  It then subtracts $\mu^*_n$ to obtain the corresponding increase in solution quality.
This comprises one loop of the algorithm.
It iterates this loop many ($J$) times and averages the differences $\mu^*_{n+1} - \mu^*_n$ obtained from different simulated values for $y_{n+1}$ to estimate the KG acquisition function $\KG_n(x)$.   
As $J$ grows large, this estimate converges to $\KG_n(x)$.

In principle, this algorithm can be used to evaluate $\KG_n(x)$ within a derivative-free simulation-based optimization method to optimize the KG acquisition function.  However, optimizing noisy simulation-based functions without access to derivatives is challenging.  \cite{frazier2009knowledge} proposed discretizing $A$ and calculating \eqref{eq:KG} exactly using properties of the normal distribution.  This works well for low-dimensional problems but becomes computationally burdensome in higher dimensions.  

\begin{algorithmic}
\begin{algorithm}
\STATE Let $\mu^{*}_n = \max_{x'} \mu_{n}(x')$.
\STATE (When calculating $\mu^*_n$ and $\mu^*_{n+1}$ below, use a nonlinear optimization method like L-BFGS.)
\FOR{$j$ = 1 to $J$:}
\STATE Generate $y_{n+1} \sim \mathrm{Normal}(\mu_n(x),\sigma^2_n(x))$.
(Equivalently, $Z\sim\mathrm{Normal}(0,1)$ and $y_{n+1} = \mu_n(x) + \sigma_n(x) Z$.)
\STATE Set $\mu_{n+1}(x';x,y_{n+1})$ to the posterior mean at $x'$ via \eqref{eq:posterior} with $(x,y_{n+1})$ as the last observation.
\STATE $\mu^*_{n+1} = \max_{x'} \mu_{n+1}(x';x,y_{n+1})$.
\STATE $\Delta^{(j)} = \mu^*_{n+1} - \mu^*_n$.
\ENDFOR
\STATE Estimate $\KG_n(x)$ by $\frac1J \sum_{j=1}^J \Delta^{(j)}$.
\caption{
Simulation-based computation of the knowledge-gradient factor $\KG_n(x)$.
\label{alg:KG}
}
\end{algorithm}
\end{algorithmic}

Overcoming this challenge with dimensionality,
\cite{wu2016parallel} proposed\footnote{This approach was proposed in the context of BayesOpt with parallel function evaluations, but can also be used in the setting considered here in which we perform one evaluation at a time.} a substantially more efficient and scalable approach, based on multi-start stochastic gradient ascent.
Stochastic gradient ascent \citep{Ro51,blum1954multidimensional} is an algorithm for finding local optima of functions used such unbiased gradient estimates widely used in machine learning \citep{bottou2012stochastic}. Multistart stochastic gradient ascent \citep{marti2016multi} runs multiple instances of stochastic gradient ascent from different starting points and selects the best local optimum found as an approximate global optimum.

We summarize this approach for maximizing the KG acquisition function in Algorithm~\ref{alg:sgd}.
The algorithm iterates over starting points, indexed by $r$, and for each maintains a sequence of iterates $x_t^{(r)}$, indexed by $t$, that converges to a local optimum of the KG acquisition function.
The inner loop over $t$ relies on a {\it stochastic gradient} $G$, which is a random variable whose expected value is equal to the gradient of the KG acquisition function with respect to where we sample, evaluated at the current iterate $x_{t-1}^{(r)}$.  We obtain the next iterate by taking a step in the direction of the stochastic gradient $G$.  The size of this step is determined by the magnitude of $G$ and a decreasing step size $\alpha_t$.  Once stochastic gradient ascent has run for $T$ iterations for each start, Algorithm~\ref{alg:sgd} uses simulation (Algorithm~\ref{alg:KG}) to evaluate the KG acquisition function for the final point obtained from each starting point, and selects the best one.

\begin{algorithmic}
\begin{algorithm}
\FOR{$r$ = 1 to $R$}
\STATE Choose $x_0^{(r)}$ uniformly at random from $A$.
\FOR{$t$ = 1 to $T$}
\STATE Let $G$ be the stochastic gradient estimate of $\nabla \KG_n(x_{t-1}^{(r)})$ from Algorithm~\ref{alg:grad-KG}.
\STATE Let $\alpha_t = a / (a+t)$.
\STATE $x_{t}^{(r)} = x_{t-1}^{(r)} + \alpha_t G$.
\ENDFOR
\STATE Estimate $\KG_n(x_T^{(r)})$ using Algorithm~\ref{alg:KG} and $J$ replications.
\ENDFOR
\STATE Return the $x_T^{(r)}$ with the largest estimated value of $\KG_n(x_T^{(r)})$.
\caption{
Efficient method for finding $x$ with the largest $\KG_n(x)$, based on 
multistart stochastic gradient ascent.  Takes as input a number of starts $R$, a number of iterations $T$ for each pass of stochastic gradient ascent, a parameter $a$ used to define a stepsize sequence, and a number of replications $J$.  Suggested input parameters: $R=10$, $T=10^2$, $a=4$, $J=10^3$.
\label{alg:sgd}
}
\end{algorithm}
\end{algorithmic}

This stochastic gradient $G$ used by the inner loop of Algorithm~\ref{alg:sgd} is calculated via Algorithm~\ref{alg:grad-KG}. 
This algorithm is based on the idea that we can exchange gradient and expectation (under sufficient regularity conditions) to write,
\begin{equation*}
\nabla \KG_n(x) = \nabla E_n\left[ \mu^*_{n+1} - \mu^*_n | x_{n+1} = x\right].
= E_n\left[ \nabla \mu^*_{n+1} | x_{n+1} = x\right],
\end{equation*}
where we have noted that $\mu^*_n$ does not depend on $x$.
This approach is called {\it infinitesimal perturbation analysis} \citep{ho1983infinitesimal}.
Thus, to construct a stochastic gradient it is sufficient to sample $\nabla \mu^*_{n+1}$.
In other words, imagine first sampling $Z$ in the inner loop in Algorithm~\ref{alg:KG}, and then holding $Z$ fixed while calculating the gradient of $\mu_{n+1}^*$ with respect to $x$.
To calculate this gradient, see that $\mu_{n+1}^*$ is a maximum over $x'$ of $\mu_{n+1}(x'; x, y_{n+1}) = \mu_{n+1}(x'; x, \mu_n(x) + \sigma_n(x) Z)$.  This is a maximum over collection of functions of $x$.
The envelope theorem \citep{milgrom2002envelope} tells us (under sufficient regularity conditions) that the gradient with respect to $x$ of a maximum of a collection of functions of $x$ is given simply by first finding the maximum in this collection, and then differentiating this single function with respect to $x$.
In our setting, we apply this by letting $\widehat{x^*}$ be the $x'$ maximizing
$\mu_{n+1}(x'; x, \mu_n(x) + \sigma_n(x) Z)$, and then calculating the gradient of 
$\mu_{n+1}(\widehat{x^*}; x, \mu_n(x) + \sigma_n(x) Z)$ with respect to $x$ while holding $\widehat{x^*}$ fixed.
In other words,
\begin{equation*}
\nabla 
\max_{x'} \mu_{n+1}(x'; x, \mu_n(x) + \sigma_n(x) Z)
= \nabla \mu_{n+1}(\widehat{x^*}; x, \mu_n(x) + \sigma_n(x) Z),
\end{equation*}
where we remind the reader that $\nabla$ refers to taking the gradient with respect to $x$, here and throughout.
This is summarized in Algorithm~\ref{alg:grad-KG}.

\begin{algorithmic}
\begin{algorithm}
\FOR{$j$ = 1 to $J$}
\STATE Generate $Z\sim\mathrm{Normal}(0,1)$
\STATE $y_{n+1} = \mu_n(x) + \sigma_n(x) Z$.
\STATE Let $\mu_{n+1}(x';x,y_{n+1})=\mu_{n+1}(x';x,\mu_n(x)+\sigma_n(x)Z)$
be the posterior mean at $x'$ computed via \eqref{eq:posterior} with $(x,y_{n+1})$ as the last observation. 
\STATE Solve $\max_{x'} \mu_{n+1}(x';x,y_{n+1})$, e.g., using L-BFGS. Let $\widehat{x^*}$ be the maximizing $x'$.
\STATE Let $G^{(j)}$ be the gradient of $\mu_{n+1}(\widehat{x^*};x,\mu_n(x)+\sigma_n(x)Z)$ with respect to $x$, holding $\widehat{x^*}$ fixed.
\ENDFOR
\STATE Estimate $\nabla\KG_n(x)$ by $G = \frac1J \sum_{j=1}^J G^{(j)}$.
\caption{
Simulation of unbiased stochastic gradients $G$ with $E[G] = \nabla \KG_n(x)$.  This stochastic gradient can then be used within stochastic gradient ascent to optimize the KG acquisition function.
\label{alg:grad-KG}
}
\end{algorithm}
\end{algorithmic}

Unlike expected improvement, which only considers the posterior at the point sampled, 
the KG acquisition considers the posterior over $f$'s full domain, and how the sample will change that posterior.
KG places a positive value on measurements that cause the maximum of the posterior mean to improve, even if the value at the sampled point is not better than the previous best point.  This provides a small performance benefit in the standard BayesOpt problem with noise-free evaluations \citep{frazier2009knowledge}, and provides substantial performance improvements in problems with noise, multi-fidelity observations, derivative observations, the need to integrate over environmental conditions, and other more exotic problem features (Section~\ref{sec:exotic}).  In these alternate problems, the value of sampling comes not through an improvement in the best solution {\it at the sampled point}, but through an improvement in the maximum of the posterior mean across feasible solutions.  For example, a derivative observation may show that the function is increasing in a particular direction in the vicinity of the sampled point.  This may cause the maximum of the posterior mean to be substantially larger than the previous maximum, even if the function value at the sampled point is worse than the best previously sampled point.  When such phenomenon are first-order, KG tends to significantly outperform EI \citep{wu2017gradient,poloczek2017multi,wu2016parallel,toscano2018bayesian}.

\subsection{Entropy Search and Predictive Entropy Search}
\label{sec:EntropySearch}
The entropy search (ES) \citep{hennig2012entropy} acquisition function values the information we have about the location of the global maximum according to its differential entropy.  
ES seeks the point to evaluate that causes the largest decrease in differential entropy.  
(Recall from, e.g., \cite{cover2012elements},
that the differential entropy of a continuous probability distribution $p(x)$ is $\int p(x) \log(p(x))\, dx$, and that smaller differential entropy indicates less uncertainty.)
Predictive entropy search (PES) \citep{hernandez2014predictive} seeks the same point, but uses a reformulation of the entropy reduction objective based on mutual information.  Exact calculations of PES and ES would give equivalent acquisition functions, but exact calculation is not typically possible, and so the difference in computational techniques used to approximate the PES and ES acquisition functions creates practical differences in the sampling decisions that result from the two approaches.  We first discuss ES and then PES.

Let $\xstar$ be the global optimum of $f$.  The posterior distribution on $f$ at time $n$ induces a probability distribution for $\xstar$.  Indeed, if the domain $A$ were finite, then we could represent $f$ over its domain by a vector $(f(x) : x \in A)$, and $\xstar$ would correspond to the largest element in this vector.  The distribution of this vector under the time-$n$ posterior distribution would be multivariate normal, and this multivariate normal distribution would imply the distribution of $\xstar$.  When $A$ is continuous, the same ideas apply, where $\xstar$ is a random variable whose distribution is implied by the Gaussian process posterior on $f$.

With this understanding, we represent the entropy of the time-$n$ posterior distribution on $\xstar$ with the notation 
$H(P_n(\xstar))$.  Similarly, $H(P_n(\xstar | x,f(x)))$ represents the entropy of what the time-$n+1$ posterior distribution on $\xstar$ will be if we observe at $x$ and see $f(x)$.  This quantity depends on the value of $f(x)$ observed.
Then, the entropy reduction due to sampling $x$ can be written,
\begin{equation}
\ES_n(x) = H(P_n(\xstar)) - E_{f(x)}\left[H(P_n(\xstar | f(x)))\right].
\label{eq:ES}
\end{equation}
In the second term, the subscript in the outer expectation indicates that we take the expectation over $f(x)$.  Equivalently, this can be written $\int \varphi(y; \mu_n(x), \sigma_n^2(x)) H(P_n(\xstar | f(x)=y))\, dy$ where $\varphi(y; \mu_n(x), \sigma_n^2(x))$ is the normal density with mean $\mu_n(x)$ and variance $\sigma_n^2(x)$.

Like KG, ES and PES below are influenced by how the measurement changes the posterior over the whole domain, and not just on whether it improves over an incumbent solution at the point sampled.  This is useful when deciding where to sample in exotic problems, and it is here that ES and PES can provide substantial value relative to EI.

While ES can be computed and optimized approximately \citep{hennig2012entropy}, doing so is challenging because
(a) the entropy of the maximizer of a Gaussian process is not available in closed form; (b) we must calculate this entropy for a large number of $y$ to approximate the expectation in \eqref{eq:ES}; and (c) we must then optimize this hard-to-evaluate function.  Unlike KG, there is no known method for computing stochastic gradients that would simplify this optimization.

PES offers an alternate approach for computing \eqref{eq:ES}.  This approach notes that the reduction in the entropy of $\xstar$ due to measuring $f(x)$ is equal to the mutual information between $f(x)$ and $\xstar$, which is in turn equal to the reduction in the entropy of $f(x)$ due to measuring $\xstar$.  This equivalence gives the expression
\begin{equation} 
\PES_n(x) = 
\ES_n(x) = H(P_n(f(x))) - E_{\xstar}\left[H(P_n(f(x) | \xstar))\right]
\end{equation} 
Here, the subscript in the expectation in the second term indicates that the expectation is taken over $\xstar$.

Unlike ES, the first term in the PES acquisition function, $H(P_n(f(x)))$, can be computed in closed form.  The second term must still be approximated: \cite{hernandez2014predictive} provides a method for sampling $\xstar$ from the posterior distribution, and a method for approximating $H(P_n(f(x) | \xstar))$ using expectation propagation \citep{minka2001family}.  This evaluation method may then be optimized by a method for derivative-free optimization via simulation.

\subsection{Multi-Step Optimal Acquisition Functions}
\label{sec:multi-step}
We can view the act of solving problem \eqref{eq:obj} as a sequential decision-making problem
\citep{GinsbourgerRiche2010,Frazier2012},
in which we sequentially choose $x_n$, and observe $y_n = f(x_n)$, with the choice of $x_n$ depending on all past observations.
At the end of these observations, we then receive a reward that might be equal to the value of the best point observed, $\max_{m \le N} f(x_m)$, as it was in the analysis of EI, or could be equal to the value $f(\widehat{\xstar})$ of the objective at some new point $\widehat{\xstar}$ chosen based on these observations as in the analysis of KG, or it could be the entropy of the posterior distribution on $\xstar$ as in ES or PES.

By construction, the EI, KG, ES, and PES acquisition functions are optimal when $N=n+1$, in the sense of maximizing the expected reward under the posterior.  However, they are no longer obviously optimal when $N>n+1$.  In principle, it is possible to compute a {\it multi-step optimal} acquisition function that would maximize expected reward for general $N$ via stochastic dynamic programming \citep{DyYu79}, but the so-called curse of dimensionality \citep{Po07} makes it extremely challenging to compute this multi-step optimal acquisition function in practice.  

Nevertheless, the literature has recently begun to deploy approximate methods for computing this solution, with attempts including \cite{lam2016bayesian,GinsbourgerRiche2010,gonzalez2016glasses}.  These methods do not yet seem to be in a state where they can be deployed broadly for practical problems, because the error and extra cost introduced in solving stochastic dynamic programming problems approximately often overwhelms the benefit that considering multiple steps provides. Nevertheless, given concurrent advances in reinforcement learning and approximate dynamic programming, this represents a promising and exciting direction for Bayesian optimization.

In addition, there are other problem settings closely related to the one most commonly considered by Bayesian optimization where it is possible to compute multi-step optimal algorithms.  For example, \cite{cashore2016multi} and \cite{xie2013sequential} use problem structure to efficiently compute multi-step optimal algorithms for certain classes of Bayesian feasibility determination problems, where we wish to sample efficiently to determine whether $f(x)$ is above or below a threshold for each $x$.  Similarly, \cite{waeber2013bisection}, building on \cite{jedynak2012twenty}, computes the multi-step optimal algorithm for a one-dimensional stochastic root-finding problem with an entropy objective.
While these optimal multi-step methods are only directly applicable to very specific settings, 
they offer an opportunity to study the improvement possible more generally by going from one-step optimal to multi-step optimal.
Surprisingly, in these settings, existing acquisition functions perform almost as well as the multi-step optimal algorithm.   For example, experiments conducted in \cite{cashore2016multi} show the KG acquisition function is within 98\% of optimal in the problems computed there, and \cite{waeber2013bisection} shows that the entropy search acquisition function is multi-step optimal in the setting considered there.  Generalizing from these results, it could be that the one-step acquisition functions are close enough to optimal that further improvement is not practically meaningful, or it could be that multi-step optimal algorithms will provide substantially better performance in yet-to-be-identified practically important settings.

\section{Exotic Bayesian Optimization} 
\label{sec:exotic}

Above we described methodology for solving the ``standard'' Bayesian optimization problem described in Section~\ref{sec:intro}.  This problem assumed
a feasible set in which membership is easy to evaluate, such as a hyperrectangle or simplex;
a lack of derivative information;
and noise-free evaluations.

While there are quite a few applied problems that meet all of the assumptions of the standard problem, there are even more where one or more of these assumptions are broken.  We call these  ``exotic'' problems.  Here, we describe some prominent examples and give references for more detailed reading.
(Although we discuss noisy evaluations in this section on exotic problems, they are substantially less exotic than the others considered, and are often considered to be part of the standard problem.) 

\paragraph{Noisy Evaluations}
GP regression can be extended naturally to observations with independent normally distributed noise of known variance \citep{RaWi06}.  This adds a diagonal term with entries equal to the variance of the noise to the covariance matrices in \eqref{eq:posterior}. In practice, this variance is not known, and so the most common approach is to assume that the noise is of common variance and to include this variance as a hyperparameter.  It is also possible to perform inference assuming that the variance changes with the domain, by modeling the log of the variance with a second Gaussian process \citep{kersting2007most}.  

The KG, ES, and PES acquisition functions apply directly in the setting with noise and they retain their one-step optimality properties.  One simply uses the posterior mean of the Gaussian process that includes noise.  

Direct use of the EI acquisition function presents conceptual challenges, however, since the ``improvement'' that results from a function value is no longer easily defined, and $f(x)$ in \eqref{eq:EI-theory} is no longer observed.  Authors have employed a variety of heuristic approaches, substituting different normal distributions for the distribution of $f(x)$ in \eqref{eq:EI-theory}, and typically using the maximum of the posterior mean at the previously evaluated points in place of $f^*_n$.  Popular substitutes for the distribution of $f(x)$ include the distribution of $\mu_{n+1}(x)$, the distribution of $y_{n+1}$, and continuing to use the distribution of $f(x)$ even though it is not observed.  Because of these approximations, KG can outperform EI substantially in problems with substantial noise \citep{wu2016parallel,frazier2009knowledge}.

As an alternative approach to applying EI when measurements are noisy, \cite{ScottFrazierPowell2011} considers noisy evaluations under the restriction made in the derivation of EI: that the reported solution needs to be a previously reported point.  It then finds the one-step optimal place to sample under this assumption.  Its analysis is similar to that used to derive the KG policy, except that we restrict $\widehat{x_*}$ to those points that have been evaluated. 

Indeed, if we were to report a final solution after $n$ measurements, it would be the point among $x_{1:n}$ with the largest value of $\mu_n(x)$, and it would have conditional expected value 
$\mu^{**}_n = \max_{i=1,\ldots,n} \mu_n(x_i)$.  If we were to take one more sample at $x_{n+1} = x$, it would have conditional expected value under the new posterior of $\mu^{**}_{n+1} = \max_{i=1,\ldots,n+1} \mu_{n+1}(x_i)$.  Taking the expected value of the difference, the value of sampling at $x$ is 
\begin{equation}
E_n\left[ \mu^{**}_{n+1} - \mu^{**}_n | x_{n+1} = x\right].
\label{eq:KGCP}
\end{equation}
Unlike the case with noise-free evaluations, this sample may cause $\mu_{n+1}(x_i)$ to differ from $\mu_n(x_i)$ for $i\le n$, necessitating a more complex calculation than in the noise-free setting (but a simpler calculation than for the KG policy).  A procedure for calculating this quantity and its derivative is given in \cite{ScottFrazierPowell2011}.
While we can view this acquisition function as an approximation to the KG acquisition function as \cite{ScottFrazierPowell2011} does (they call it the KGCP acquisition function), we argue here that it is the most natural generalization of EI's assumptions to the case with noisy measurements.

\paragraph{Parallel Evaluations}
Performing evaluations in parallel using multiple computing resources allow obtaining multiple function evaluations in the time that would ordinarily be required to obtain just one with sequential evaluations.  For this reason, parallel function evaluations is a conceptually appealing way to solve optimization problems in less time.  
EI, KG, ES, and PES can all be extended in a straightforward way to allow parallel function evaluations.  For example, EI becomes 
\begin{equation}
\label{eq:parallel-EI}
\EI_n(x^{(1:q)}) = E_n\left[[\max_{i=1,\ldots,q} f(x^{(i)}) - f^*_n]^+\right],
\end{equation}
where $x^{(1:q)} = (x^{(1)},\ldots, x^{(q)})$ is a collection of points at which we are proposing to evaluate \citep{GiLeCa08}.  Parallel EI (also called {\it multipoints EI} by \cite{GiLeCa08}) then proposes to evaluate the set of points that jointly maximize this criteria.  This approach can also be used asynchronously, where we hold fixed those $x^{(i)}$ currently being evaluated and we allocate our idle computational resources by optimizing over their corresponding $x^{(j)}$.

Parallel EI \eqref{eq:parallel-EI} and other parallel acquisition functions are more challenging to optimize than their original sequential versions from Section~\ref{sec:acquisition}.  One innovation is the Constant Liar approximation to the parallel EI acquisition function \citep{GiLeCa10}, which chooses $x^{(i)}$ sequentially by assuming that $f(x^{(j)})$ for $j<i$ have been already observed, and have values equal to a constant (usually the expected value of $f(x^{(j)}))$ under the posterior.  This substantially speeds up computation.  
Expanding on this, \cite{wang2016parallel} showed that infinitesimal perturbation analysis can produce random stochastic gradients that are unbiased estimates of $\nabla \EI_n(x^{(1:q)})$, which can then be used in multistart stochastic gradient ascent to optimize \eqref{eq:parallel-EI}.  This method has been used to implement the parallel EI procedure for as many as $q=128$ parallel evaluations.
Computational methods for parallel KG were developed by \cite{wu2016parallel}, and are implemented in the Cornell MOE software package discussed in Section~\ref{sec:software}.  That article follows the stochastic gradient ascent approach described above in Section~\ref{sec:KG}, which generalizes well to the parallel setting.

\paragraph{Constraints}
In the problem posed in Section~\ref{sec:intro}, we assumed that the feasible set was a simple one in which it was easy to assess membership.  The literature has also considered the more general problem,
\begin{align*}
&\max_x f(x) \\
&\text{subject to}\ g_i(x) \ge 0, \quad i=1,\ldots,I,
\end{align*}
where the $g_i$ are as expensive to evaluate as $f$.  EI generalizes naturally to this setting when $f$ and $g_i$ can be evaluated without noise: improvement results when the evaluated $x$ is feasible ($g_i(x) \ge 0$ for all $x$) and $f(x)$ is better than the best previously evaluated feasible point.  
This was proposed in Section 4 of \cite{schonlau1998global}
and studied independently by \cite{gardner2014bayesian}.
PES has also been studied for this setting \citep{hernandez2015predictive}.

\paragraph{Multi-Fidelity and Multi-Information Source Evaluations}
\nocite{poloczek2017multi}

In multi-fidelity optimization, rather than a single objective $f$, we have a collection of information sources $f(x,s)$ indexed by $s$.  Here, $s$ controls the ``fidelity'', with lower $s$ giving higher fidelity, and $f(x,0)$ corresponding to the original objective.  Increasing the fidelity (decreasing $s$) gives a more accurate estimate of $f(x,0)$, but at a higher cost $c(s)$.  
For example, $x$ might describe the design of an engineering system, and $s$ the size of a mesh used in solving a partial differential equation that models the system.  Or, $s$ might describe the time horizon used in a steady-state simulation.  Authors have also recently considered optimization of neural networks, where $s$ indexes the number of iterations or amount of data used in training a machine learning algorithm \citep{swersky2014freeze,klein2016fast}.
Accuracy is modeled by supposing that $f(x,s)$ is equal to $f(x,0)$ and is observed with noise whose variance $\lambda(s)$ increases with $s$, or by supposing that $f(x,s)$ provides deterministic evaluations with $f(x,s+1)-f(x,s)$ modeled by a mean $0$ Gaussian process that varies with $x$.  Both settings can be modeled via a Gaussian process on $f$, including both $x$ and $s$ as part of the modeled domain.

The overarching goal is to solve $\max_x f(x,0)$ by observing $f(x,s)$ at a sequence of points and fidelities $(x_n,s_n)$ with total cost $\sum_{n=1}^N c(s_n)$ less than some budget $B$.  Work on multi-fidelity optimization includes \cite{HuAlNoMi06,SobesterLearyKeane2004,forrester2007multi,mcleod2017practical,kandasamy2016gaussian}.  

In the more general problem of multi-information source optimization, we relax the assumption that the $f(\cdot,s)$ are ordered by $s$ in terms of accuracy and cost.  Instead, we simply have a function $f$ taking a design input $x$ and an information source input $s$, with $f(x,0)$ being the objective, and $f(x,s)$ for $s\ne0$ being observable with different biases relative to the objective, different amounts of noise, and different costs.

For example, $x$ might represent the design of an aircraft's wing, $f(x,0)$ the predicted performance of the wing under an accurate but slow simulator, and $f(x,s)$ for $s=1,2$ representing predicted performance under two inexpensive approximate simulators making different assumptions.  It may be that $f(x,1)$ is accurate for some regions of the search space and substantially biased in others, with $f(x,2)$ being accurate in other regions.  In this setting, the relative accuracy of $f(x,1)$ vs. $f(x,2)$ depends on $x$.  Work on multi-information source optimization includes \cite{lam2015multifidelity,poloczek2017multi}.

EI is difficult to apply directly in these problems because evaluating $f(x,s)$ for $s\ne 0$ never provides an improvement in the best objective function value seen, $\max \{ f(x_n,0) : s_n = 0 \}$.  Thus, a direct translation of EI to this setting causes $EI=0$ for $s\ne 0$, leading to measurement of only the highest fidelity.  For this reason, the EI-based method from \cite{lam2015multifidelity}
uses EI to select $x_n$ assuming that $f(x,0)$ will be observed (even if it will not), and uses a separate procedure to select $s$.  KG, ES, and PES can be applied directly to these problems, as in \cite{poloczek2017multi}. 

\paragraph{Random Environmental Conditions and Multi-Task Bayesian Optimization}
Closely related to multi-information source optimization is the pair of problems
\begin{align*}
&\max_x \int f(x,w) p(w)\,dw,\\
&\max_x \sum_w f(x,w) p(w),
\end{align*}
where $f$ is expensive to evaluate.  
These problems appear in the literature with a variety of names: optimization with random environmental conditions \citep{chang2001design} in statistics, multi-task Bayesian optimization \citep{swersky2013multi} in machine learning, along with optimization of integrated response functions \citep{williams2000sequential} and optimization with expensive integrands \citep{toscano2018bayesian}.

Rather than taking the objective $\int f(x,w) p(w)\,dw$ as our unit of evaluation, a natural approach is to evaluate $f(x,w)$ at a small number of $w$ at an $x$ of interest.  This gives partial information about the objective at $x$.  Based on this information, one can explore a different $x$, or resolve the current $x$ with more precision.  Moreover, by leveraging observations at $w$ for a nearby $x$, one may already have substantial information about a particular $f(x,w)$ reducing the need to evaluate it.  Methods that act on this intuition can substantially outperform methods that simply evaluate the full objective in each evaluation via numerical quadrature or a full sum \citep{toscano2018bayesian}.

This pair of problems arise in the design of engineering systems and biomedical problems, such as joint replacements \citep{chang2001design} and cardiovascular bypass grafts \citep{xie2012optimization}, where $f(x,w)$ is the performance of design $x$ under environmental condition $w$ as evaluated by some expensive-to-evaluate computational model, $p(w)$ is some simple function (e.g., the normal density) describing the frequency with which condition $w$ occurs, and our goal is to optimize average performance.
It also arises in machine learning, in optimizing cross-validation performance.  Here, we divide our data into chunks or ``folds'' indexed by $w$, and $f(x,w)$ is the test performance on fold $w$ of a machine learning model trained without data from this fold.

Methods in this area include KG for noise-free \citep{xie2012optimization} and general problems \citep{toscano2018bayesian}, PES \citep{swersky2013multi}, and modifications of EI \citep{groot2010bayesian,williams2000sequential}.
As in multi-information source optimization, the unmodified EI acquisition function is inappropriate here because observing $f(x,w)$ does not provide an observation of the objective (unless all $w'\ne w$ have already been observed at that $x$) nor a strictly positive improvement.
Thus, \cite{groot2010bayesian} and \cite{williams2000sequential} use EI to choose $x$ as if it we did observe the objective, and then use a separate strategy for choosing $w$.

\paragraph{Derivative Observations}
Finally, we discuss optimization with derivatives.  Observations of $\nabla f(x)$, optionally with normally distributed noise, may be incorporated directly into GP regression \citep[Sect 9.4]{RaWi06}.  \cite{Li08} proposed using gradient information in this way in Bayesian optimization, together with the EI acquisition function, showing an improvement over BFGS \citep{liu1989limited}.  EI is unchanged by a proposed observation of $\nabla f(x)$ in addition to $f(x)$ as compared to its value when observing $f(x)$ alone.  (Though, if previous derivative observations have contributed to the time $n$ posterior, then that time-$n$ posterior will differ from what it would be if we had observed only $f(x)$.) Thus, EI does not take advantage of the availability of derivative information to, for example, evaluate at points far away from previously evaluated ones where derivative information would be particularly useful.  A KG method alleviating this problem was proposed by \cite{wu2017gradient}.  In other related work in this area, \cite{osborne2009gaussian} proposed using gradient information to improve conditioning of the covariance matrix in GP regression, and \cite{ahmed2016we} proposed a method for choosing a single directional derivative to retain when observing gradients to improve the computational tractability of GP inference.


\section{Software}
\label{sec:software}
There are a variety of codes for Bayesian optimization and Gaussian process regression.
Several of these Gaussian process regression and Bayesian optimization packages are developed together, with the Bayesian optimization package making use of the Gaussian process regression package.  Other packages are standalone, providing only either Gaussian process regression support or Bayesian optimization support.
We list here several of the most prominent packages, along with URLs that are current as of June 2018.
\begin{itemize}
\item DiceKriging and DiceOptim are packages for Gaussian process regression and Bayesian optimization respectively, written in R.  They are described in detail in \cite{JSSv051i01} and are available from CRAN via \url{https://cran.r-project.org/web/packages/DiceOptim/index.html}.
\item GPyOpt (\url{https://github.com/SheffieldML/GPyOpt}) is a python Bayesian optimization library built on top of the Gaussian process regression library GPy (\url{https://sheffieldml.github.io/GPy/}) both written and maintained by the machine learning group at Sheffield University.
\item Metrics Optimization Engine (MOE, \url{https://github.com/Yelp/MOE}) is a Bayesian optimization library in C++ with a python wrapper that supports GPU-based computations for improved speed.  It was developed at Yelp by the founders of the Bayesian optimization startup, SigOpt (\url{http://sigopt.com}).  Cornell MOE (\url{https://github.com/wujian16/Cornell-MOE}) is built on MOE with changes that make it easier to install, and support for parallel and derivative-enabled knowledge-gradient algorithms.
\item Spearmint (\url{https://github.com/HIPS/Spearmint}), with an older version under a different license available at \url{https://github.com/JasperSnoek/spearmint}, is a python Bayesian optimization library.  Spearmint was written by the founders of the Bayesian optimization startup Whetlab, which was acquired by Twitter in 2015 \cite{whetlab}.
\item DACE (Design and Analysis of Computer Experiments) is a Gaussian process regression library written in MATLAB, available at \url{http://www2.imm.dtu.dk/projects/dace/}.  Although it was last updated in 2002, it remains widely used.
\item GPFlow (\url{https://github.com/GPflow/GPflow}) and
GPyTorch (\url{https://github.com/cornellius-gp/gpytorch})
are python Gaussian process regression library built on top of Tensorflow (\url{https://www.tensorflow.org/}) 
and PyTorch (\url{https://pytorch.org/}) respectively.
\item laGP (\url{https://cran.r-project.org/web/packages/laGP/index.html}) is an R package for Gaussian process regression and Bayesian optimization with support for inequality constraints.
\end{itemize}

\section{Conclusion and Research Directions}
\label{sec:conclusion}
We have introduced Bayesian optimization, first discussing GP regression, then the expected improvement, knowledge gradient, entropy search, and predictive entropy search acquisition functions.  We then discussed a variety of exotic Bayesian optimization problems: those with noisy measurements; parallel evaluations; constraints; multiple fidelities and multiple information sources; random environmental conditions and multi-task BayesOpt; and derivative observations.

Many research directions present themselves in this exciting field.
First, there is substantial room for developing a deeper theoretical understanding of Bayesian optimization.  As described in Section~\ref{sec:multi-step}, settings where we can compute multi-step optimal algorithms are extremely limited.  Moreover, while the acquisition functions we currently use in practice seem to perform almost as well as optimal multi-step algorithms when we can compute them, we do not currently have finite-time bounds that explain their near-optimal empirical performance, nor do we know whether multi-step optimal algorithms can provide substantial practical benefit in yet-to-be-understood settings.  Even in the asymptotic regime, relatively little is known about rates of convergence for Bayesian optimization algorithms: while \cite{bull2011convergence} establishes a rate of convergence for expected improvement when it is combined with periodic uniform sampling, it is unknown whether removing uniform sampling results in the same or different rate.

Second, there is room to build Bayesian optimization methods that leverage novel statistical approaches.  Gaussian processes (or variants thereof such as \cite{snoek2014input} and \cite{kersting2007most}) are used in most work on Bayesian optimization, but it seems likely that classes of problems exist where the objective could be better modeled through other approaches.  It is both of interest to develop new statistical models that are broadly useful, and to develop models that are specifically designed for applications of interest.

Third, developing Bayesian optimization methods that work well in high dimensions is of great practical and theoretical interest.  Directions for research include developing statistical methods that identify and leverage structure present in high-dimensional objectives arising in practice, which has been pursued by recent work including \cite{wang2013bayesian,wang2016bayesian,kandasamy2015high}.  See also \cite{shan2010survey}.
It is also possible that new acquisition functions may provide substantial value in high dimensional problems. 

Fourth, it is of interest to develop methods that leverage exotic problem structure unconsidered by today's methods, in the spirit of the Section~\ref{sec:exotic}.  It may be particularly fruitful to combine such methodological development with applying Bayesian optimization to important real-world problems, as using methods in the real world tends to reveal unanticipated difficulties and spur creativity.  

Fifth, substantial impact in a variety of fields seems possible through application of Bayesian optimization.  One set of application areas where Bayesian optimization seems particularly well-positioned to offer impact is in chemistry, chemical engineering, materials design, and drug discovery, where practitioners undertake design efforts involving repeated physical experiments consuming years of effort and substantial monetary expense.  While there is some early work in these areas \citep{ueno2016combo,frazier2016bayesian,FrazierNegoescuPowell2011,seko2015prediction,ju2017designing} the number of researchers working in these fields aware of the power and applicability of Bayesian optimization is still relatively small.

\section*{Acknowledgments}
While writing this tutorial, the author was supported by the Air Force Office of Scientific Research and the National Science Foundation
(AFOSR FA9550-15-1-0038 and NSF CMMI-1254298, CMMI-1536895 DMR-1719875, DMR-1120296).  The author would also like to thank Roman Garnett, whose suggestions helped shape the discussion of expected improvement with noise, and several anonymous reviewers.

\bibliography{tutorial}
\bibliographystyle{apalike}

\end{document}